\pdfoutput=1

\documentclass[11pt]{article}

\usepackage[preprint]{acl}

\usepackage{times}
\usepackage{latexsym}

\usepackage[T1]{fontenc}

\usepackage[utf8]{inputenc}

\usepackage{microtype}

\usepackage{inconsolata}

\usepackage{graphicx}
\usepackage{hyperref}
\usepackage{url}
\usepackage{booktabs}
\usepackage{amsfonts}
\usepackage{multirow}
\usepackage[linesnumbered,ruled,vlined]{algorithm2e}
\usepackage{amssymb}
\usepackage{amsmath}
\usepackage{bbm}
\usepackage{subcaption}

\title{Prompt-based Pseudo-labeling Strategy for Sample-Efficient Semi-Supervised Extractive Summarization}

\author{Gaurav Sahu \& Olga Vechtomova \\
David R. Cheriton School of Computer Science\\
University of Waterloo, Canada\\
\texttt{\{gaurav.sahu\}@uwaterloo.ca} \\
\And
Issam H. Laradji \\
ServiceNow Research \\
University of British Columbia \\}

\begin{document}
\maketitle
\begin{abstract}
Semi-supervised learning (SSL) is a widely used technique in scenarios where labeled data is scarce and unlabeled data is abundant. While SSL is popular for image and text classification, it is relatively underexplored for the task of extractive text summarization.
Standard SSL methods follow a teacher-student paradigm to first train a classification model and then use the classifier's confidence values to select pseudo-labels for the subsequent training cycle; however, such classifiers are not suitable to measure the accuracy of pseudo-labels as they lack specific tuning for evaluation, which leads to confidence values that fail to capture the semantics and correctness of the generated summary.
To address this problem, we propose a prompt-based pseudo-labeling strategy with LLMs that picks unlabeled examples with more accurate pseudo-labels than using just the classifier's probability outputs.
Our approach also includes a relabeling mechanism that improves the quality of pseudo-labels.
We evaluate our method on three text summarization datasets: TweetSumm, WikiHow, and ArXiv/PubMed.
We empirically show that a prompting-based LLM that scores and generates pseudo-labels outperforms existing SSL methods on ROUGE-1, ROUGE-2, and ROUGE-L scores on all the datasets.
Furthermore, our method achieves competitive L-Eval scores (evaluation with LLaMa-3) as a fully supervised method in a data-scarce setting and outperforms fully supervised method in a data-abundant setting.

\end{abstract}

\section{Introduction}

\begin{figure}[t]
	\centering
	\includegraphics[width=\linewidth]{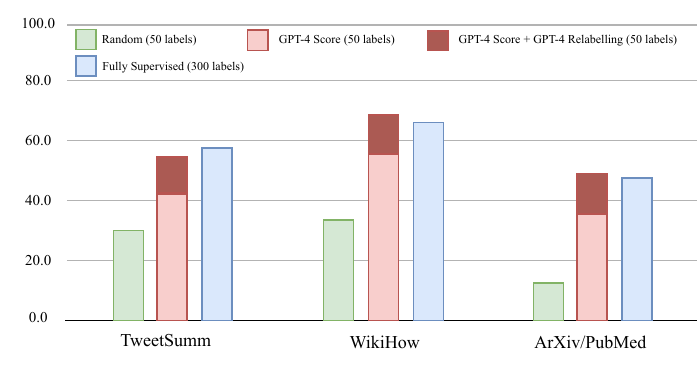}  
	\caption{\textbf{L-Eval scores of semi-supervised v/s fully supervised models in a low-data setting.} The proposed semi-supervised method (middle column for each dataset) combines PreSumm's confidence with GPT-4's knowledge to generate better pseudo-labels.
	The proposed method performs competitively on the three summarization datasets while using 6$\times$ fewer labels than a fully supervised method (rightmost column for each dataset)}
	\label{fig:teaser}
\end{figure}

Text summarization is a crucial task in today's data-driven era.
Major enterprises store customer service chatlogs and wish to create an automated text summarization system to enhance the user experience~\citep{feigenblat-etal-2021-tweetsumm-dialog}.
With the exponential growth of digital content, the need for automated summarization techniques that swiftly process and distill information has become paramount~\citep {koupaee2018wikihow,cohan-etal-2018-discourse}.
Text summarization systems are broadly categorized into two types: abstractive, where the summaries are concise \textit{paraphrases} of the original text~\citep{abs1,abs2}, and extractive, where the summaries combine existing sentences \textit{verbatim} from the original text~\citep{radev2004mead,li2006extractive,wong2008extractive}.
While both systems have their advantages, in today's fast-paced lifestyle, extractive summarization systems have become increasingly relevant as they are fast and interpretable while preserving the source integrity and reducing potential misinterpretation.

Obtaining a large annotated extractive summarization dataset to train a fully supervised model is financially expensive and it also incurs a high cost of human labor~\citep{wang2022noise}.
Real-life scenarios often have a small labeled set alongside a large pool of unlabeled data.
Semi-supervised learning (SSL) has been successfully used in such scenarios for images and text classification~\citep{img1,img2,img3,text1,text2}, but it is relatively underexplored for extractive text summarization.
In this work, we focus on extractive text summarization using SSL techniques.

Standard deep learning-based methods reformulate extractive summarization as a binary text classification problem: given a sentence in the input document, predict its likelihood of appearing in the output summary~\citep{amini2001automatic,wong2008extractive}.
SSL techniques, in particular, follow a teacher-student paradigm to train a classifier on the limited labeled data first to generate pseudo-labels for the unlabeled set and then use the same classifier's confidence values to select the best pseudo-labels for the next training cycle.
Most recently, \citet{liu-lapata-2019-text} and \citet{zhuang2023self} employed this strategy using large language models (LLMs) like BERT~\citep{devlin2019bert} as their backbones.
While these works have shown promising results, the underlying classification model is not suitable to determine the accuracy of pseudo-labels because it lacks specific fine-tuning for measuring pseudo-label accuracy.
On the other hand, recent instruction-tuned LLMs like GPT-4~\citep{openai2023gpt} and LLaMA-2 and LLaMa-3~\citep{touvron2023llama} have been successfully used for generating pseudo-labels and evaluation for a wide range of tasks~\citep{sahu-etal-2022-data,sahu2023promptmix,dziri2023faith,mishra2023llm,liu-etal-2023-g,zhang2023xdial}.

Motivated by the success, this work uses a generative LLMs to improve the process of pseudo-labeling during the training phase of an SSL algorithm.
Scoring all the pseudo-labels with LLMs can quickly get expensive as the number of pseudo-labels increases, so we first use the classification model's confidence values to {shortlist} a fixed number of pseudo-labels.
Then, we prompt an LLM with specific instructions to score the pseudo-labels and select top-$n$ pseudo-labels to include in the subsequent training cycle.
Our method also includes a prompt-based relabeling mechanism that improves pseudo-labels' accuracy.
We evaluate our approach on three text summarization datasets: TweetSumm~\citep{feigenblat-etal-2021-tweetsumm-dialog}, WikiHow~\citep{koupaee2018wikihow}, and ArXiv/PubMed~\citep{cohan-etal-2018-discourse}.
We show that our method enhances the performance of a standard teacher-student pipeline to achieve competitive performance to that of a fully supervised model while using 6$\times$ less labeled data in a data-scarce setting and outperforms fully supervised models in data-abundant setting.
Figure~\ref{fig:teaser} further demonstrates the effectiveness of our method.

\begin{figure*}[t!]
    \centering
    \includegraphics[width=0.85\linewidth]{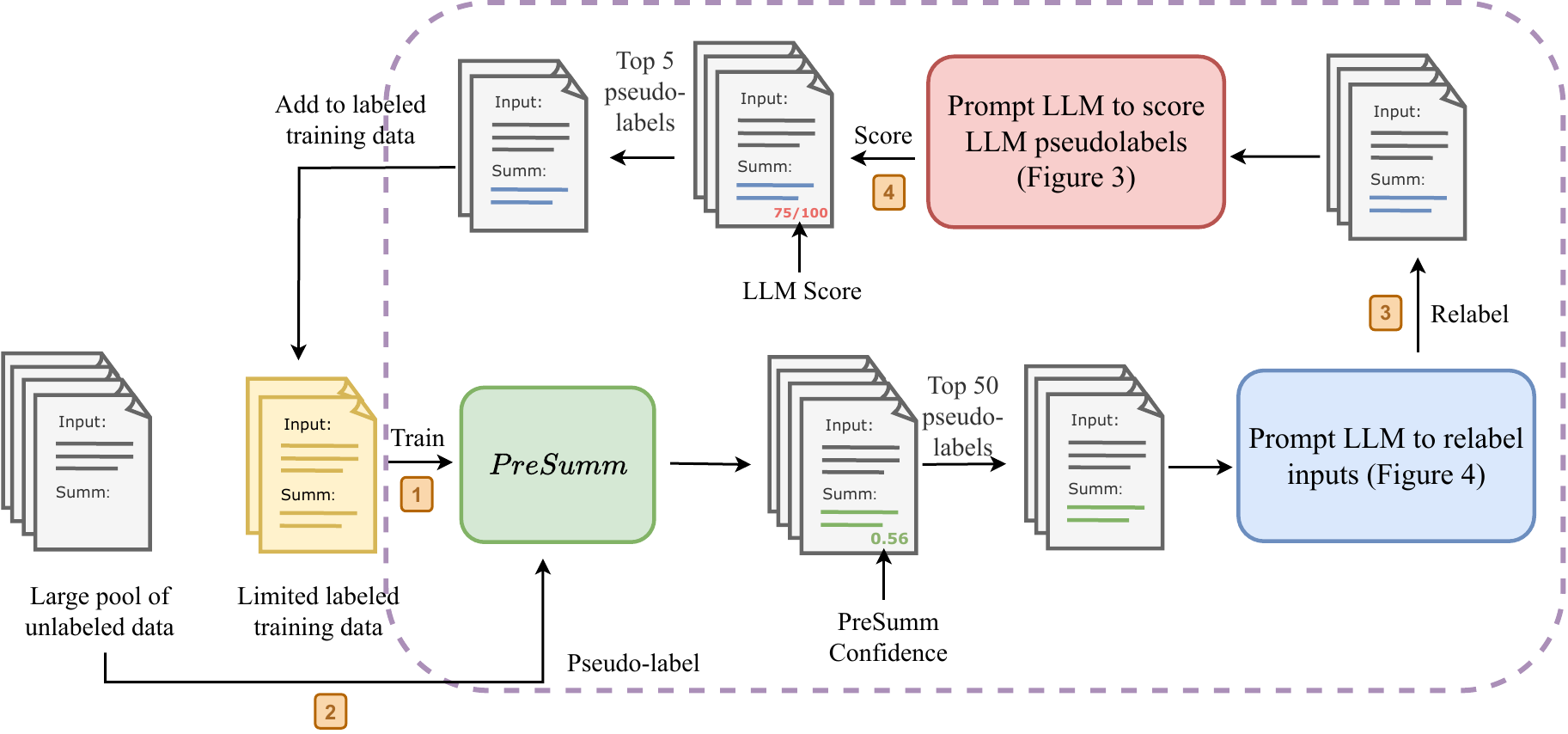}
    \caption{\textbf{Proposed method.} Our approach has four main stages. {\textbf{Stage 1:}} train a teacher model $M$ (PreSumm) on the limited labeled dataset. {\textbf{Stage 2:}} generate pseudo-labels for the unlabeled set with $M$ and shortlist 50 based on teacher confidence (see Equation~\ref{eq:conf}). \textbf{Stage 3:} prompt an off-the-shelf LLM like GPT-4 to generate an extractive summary of the shortlisted documents from Stage 2. \textbf{Stage 4:} score the pseudo-labels in Stage 3 by prompting an LLM and select the top 5. These summaries are then added to the training data for the next iteration.}
    \label{fig:fig1}
\end{figure*}

In summary, we \textbf{1)} propose a prompt-based pseudo-label selection strategy for semi-supervised extractive text summarization. \textbf{2)} We also propose a relabeling mechanism using an LLM (such as GPT-4) to generate high-quality summaries for unlabeled examples. \textbf{3)} Finally, we show that our method is more sample-efficient and outperforms prior semi-supervised methods for extractive summarization on ROUGE-1, ROUGE-2, ROUGE-L, and L-Eval (same as L-Eval but using LLaMA-3 as the LLM) scores across three benchmark datasets.

\section{Related Work}

The earliest works on extractive text summarization use unsupervised and supervised algorithms to extract sentence and word-level features.
Examples of unsupervised approaches include using graph-based ranking techniques like LexRank~\citep{graph2} and external knowledge from Wikipedia~\citep{graph1}, and fuzzy-logic systems~\citep{fuzzy1,fuzzy2}.
While these methods improve coherency and reduce redundancy, they rely on surface-level features and cannot handle dangling anaphoras.
Our method combines richer sentence representations of pretrained language models like BERT and DistilBERT~\citep{sanh2019distilbert}  with GPT-4, which has recently been shown to help in anaphora resolution~\citep{shen2023large}.

Prior supervised approaches use probabilistic support vector machines~\citep{wong2008extractive}, Bayes rule~\citep{sup1}, conditional random fields~\citep{sup2}, and more recently, pretrained language models like BERT~\citep{liu-lapata-2019-text,pilault2020extractive,zhong2020extractive} for the task.
While supervised methods provide better representation for sentence selection, they need large human-annotated datasets for training, which can be financially expensive to gather.
To address this problem, \citet{zhuang2023self} explore the semi-supervised setting and propose a novel objective function to improve the pretraining of BERT for dialog summarization.
Their method significantly improves over the traditional pretraining process; however, it lacks flexibility and is heavily tailored for dialog summarization.
Our method uses a generic prompt-based strategy with LLMs that can be easily adapted across different domains.
Additionally, our method uses less data and generates higher-quality pseudo-labels.

With the recent advancements in generative language modeling, a new body of literature using GPT-like models for summarization has emerged.
However, they have been focused primarily on abstractive setups
~\citep{dou2021gsum,wang2022salience,keswani2024abstractive}, which makes them prone to hallucination and potential misinterpretation of the original text.
\citet{zhang2023extractive} propose an extract-then-generate method where they use in-context learning to generate extractive-summary-guided abstractive summaries.
However, since they operate in a fully-supervised setting, the method suffers from scalability issues for large datasets.
\citet{mishra2023llm} propose LLM pseudo-labeling for semi-supervised dialog summarization, but our proposed method is more sample-efficient as we use fewer labeled and unlabeled examples.

\section{Methodology}
\label{sec:method}
This section describes our approach for semi-supervised extractive text summarization.
As shown in Figure~\ref{fig:fig1}, we employ a teacher-student training framework and divide our pipeline into four stages.
\textbf{First}, we train a teacher model on the available labeled examples.
\textbf{Second}, we use the trained model to generate pseudo-labels for the unlabeled set and shortlist 50 pseudo-labels based on the model's confidence.
\textbf{Third}, we prompt a generative LLM like GPT-4 to relabel the shortlisted pseudo-labels.
\textbf{Lastly}, we prompt the LLM to score the new pseudo-labels and select the top 5 to include in the next training cycle.
The following sections describe individual stages in more detail.

\begin{figure*}[t]
  \centering
  \begin{subfigure}[b]{0.4\linewidth}
    \centering
    \includegraphics[width=\linewidth]{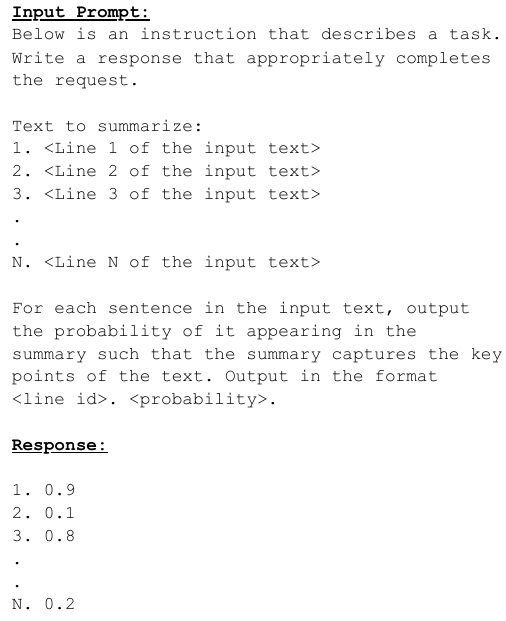}  
    \caption{\textbf{Generating pseudo-labels.} We attach a line ID to each sentence in the input document and instruct the LLM to use those line IDs in its response.}
    \label{fig:gptrelabel}
  \end{subfigure}
  \hfill
  \begin{subfigure}[b]{0.53\linewidth}
    \centering
    \includegraphics[width=\linewidth]{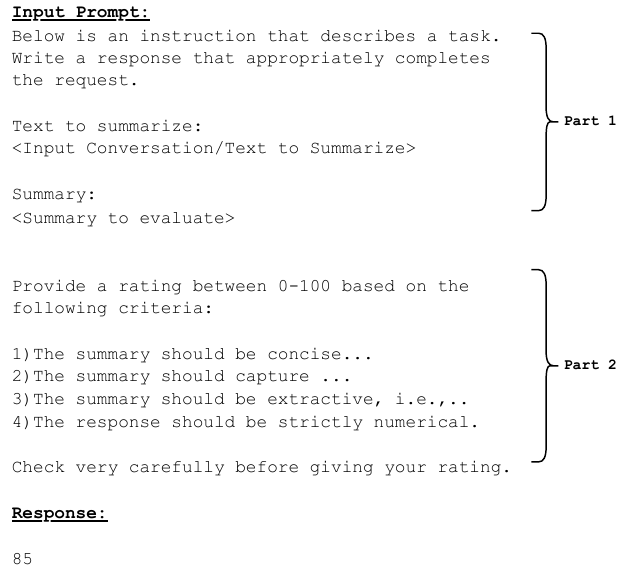}  
    \caption{\textbf{Scoring pseudo-labels.} The two-part prompt contains a text and summary pair (Part 1), and a list the evaluation criteria (Part 2). \textbf{Note:} Refer to Section~\ref{sec:score} for complete details on the evaluation criteria.}
    \label{fig:gpt4score}
  \end{subfigure}
  \caption{Different prompts used in the experiments.}
  \label{fig:whole}
\end{figure*}

\subsection{Stage 1: Training the Teacher Model}
First, we train a fully-supervised model $M$ (teacher) on the set of available labeled examples $D_l$.
We use PreSumm~\citep{liu-lapata-2019-text} as our summarization model $M$ as it has been shown to perform well for extractive summarization.
Notably, PreSumm reformulates the task of generating extractive summaries to binary classification, where, for each sentence in the input document, the model predicts if it will be present in the output summary.
Then, the model combines the top-$n$ sentences with the highest probabilities in their order of appearance in the original text to construct the extractive summary.

\subsection{Stage 2: Generating Pseudo-labels using the Teacher Model}
We use the teacher model $M$ to generate pseudo-labels for the unlabeled set $D_u$.
Next, we shortlist a subset of 50 pseudo-labels with the highest teacher confidence~\footnote{We experiment 5, 10, 20, 50, and 75 and find 50 to be optimal.}.
We compute the teacher confidence for a generated summary (a.k.a pseudo-label) as follows: let $p_{ij}$ denote the probability with which the $i$-th sentence $s_i$ in an unlabeled document $u_j$ is present in its summary $S_j$, and let $\mathbbm{1}$ denote the indicator function: $\mathbbm{1}(s_i) =
    \begin{cases}
        1, & \text{if}\ s_i \in S \\
        0, & \text{otherwise}
    \end{cases}.$
We then compute the teacher confidence for the pseudo-label $S_j$ by averaging the probabilities of selected sentences. We define the teacher confidence ($C_j$) for an input text $u_j$ as follows:

\begin{equation}
    C_j = \frac{\sum_{i=1}^{|u_j|} \mathbbm{1}(s_i) \cdot p_{ij}}{n},
\label{eq:conf}
\end{equation}

where $|u_j|$ denotes the number of sentences in the unlabeled document $u_j$ and $n$ is the number of sentences in the generated summary.

We will show in Section~\ref{sec:results} that shortlisting a subset of pseudo-labels helps make our method more sample-efficient, as we avoid relabeling a large unlabeled pool. This ultimately minimizes our LLM usage cost in the subsequent stages.

\subsection{Stage 3: LLM Relabeling of Teacher's Pseudo-labels}
After selecting the top 50 pseudo-labels using teacher confidence defined in Equation~\ref{eq:conf}, we prompt an LLM (GPT-4/LLaMA, in our case) to generate a summary for each shortlisted unlabeled example.
This effectively relabels the pseudo-label from Stage 2.
Specifically, we follow the prompt template in Figure~\ref{fig:gptrelabel} when generating summaries using GPT-4, which uses the same mechanism as PreSumm, i.e., we instruct the LLM to output probabilities for each sentence in the input document and then concatenate the top-$n$ lines in their order of appearance in the input text.

\begin{algorithm}[t]
  \small
  \SetAlgoLined
  \SetKwInOut{Input}{Input}
  \SetKwInOut{Output}{Output}
  
  \Input{ Labeled Data ${D}_l$, Unlabeled Data ${D}_u$, Number of cycles $N_{cycles}$}
  \Output{Trained summarization model $M$}
  
  \BlankLine
    $M \gets null$ \;
  \For{$i \gets 1$ \KwTo $N_{cycles}$}{
  \BlankLine
  \texttt{// Train PreSumm on labeled set}\;
    $M \gets PreSumm({D}_l)$\;
    \BlankLine
    \texttt{// Generate Pseudo-labels}\;
    $Pseudos \gets Top50({M}({D}_u))$\;
    $Pseudos \gets Top5(LLM(Pseudos))$\;
    \BlankLine
    \texttt{// Add top-5 pseudos to labeled set and update unlabeled set}\;
    $D_l \gets D_l \cup Pseudos$\;
    $D_u \gets D_u \setminus Pseudos$\;
  }
  \caption{An algorithm for prompt-based semi-supervised text summarization}
  \label{alg:ssl}
\end{algorithm}

\subsection{Stage 4: LLM Scoring of Pseudo-labels}
\label{sec:score}
In the last stage of our pipeline, we prompt LLaMA-3 as shown in Figure~\ref{fig:gpt4score} to output a rating between 0-100 for the pseudo-labels from Stage 3.
Specifically, we list the following criteria for evaluating the pseudo-labels:

\begin{itemize}
    \item The summary should be concise, ideally with less than $k$ sentences.
    \item The summary should capture the key points from the input text. Do not penalize if it doesn't include any minor additional details.
    \item The summary should be extractive, i.e., directly taken from the conversation without any modifications or removal of symbols.
    \item The response should contain only the numerical rating.
\end{itemize}

Here, $k$ is a hyperparameter chosen based on the average summary length in the dataset.
Finally, we choose the top 5 pseudo-labeled examples with the highest LLaMA scores and append them to the existing labeled set $D_l$ for the next training cycle.

We repeat Stages 1-4 $N_{cycles}$ times to improve the initial summarization model $M$ and use the model obtained after the last cycle for generating summaries for the unseen test set.
Algorithm~\ref{alg:ssl} also provides an overview of the proposed semi-supervised approach.
To note, $M(D_l)$ denotes generating summaries for $D_l$ using $M$, and $LLM(Pseudos)$ denotes the prompt-based relabeling of pseudo-labels from Stage 2.
Furthermore, $TopK(\cdot)$ denotes scoring pseudo-labels (based on teacher confidence in Stage 2 and LLM score in Stage 4) and selecting the top $K$ pseudo-labels.

\section{Experimental Setup}
This section describes our experimental setup, including the datasets, training and evaluation details, and the baselines used.

\subsection{Datasets}
We use three datasets for extractive summarization.

\noindent \textbf{1)} \textbf{TweetSumm}~\citep{feigenblat-etal-2021-tweetsumm-dialog} is a real-world customer service chat dataset. The dataset has 1100 conversations between a customer and an agent, and each conversation has three human-annotated extractive summaries.
The training set has 858 dialogs, out of which we randomly sample a subset of 50 examples as the initial training set and treat the remaining as the unlabeled set.
We use the official validation and test sets for evaluation.

\noindent \textbf{2) WikiHow} \citep{koupaee2018wikihow} contains WikiHow articles with their headlines as abstractive summaries. The dataset has over 180k articles, with around 168k training articles and 6000 test and validation articles.

\noindent\textbf{3) ArXiv/PubMed} \citep{cohan-etal-2018-discourse} is a collection of scientific articles from PubMed and ArXiv with their abstracts as their summaries.
The dataset has close to 325k articles, with nearly 300k training articles and 12.5k test and validation articles.

\begin{table}[h]
    \small 
    \centering
    \begin{tabular}{lcccc}
         \toprule
                 & TweetSumm & WikiHow & ArXiv/PubMed \\
         \midrule
         \# Train & 858 & 168,000 & 300,000 \\
         \# Valid & 100 & 6,000 & 12,500 \\
         \# Test & 100 & 6,000 & 12,500 \\
        \bottomrule          
    \end{tabular}
    \caption{Statistics of the text summarization datasets we use in our experiments.}
    \label{tab:datastats}
\end{table}

Table~\ref{tab:datastats} summarizes the dataset statistics.
Notably, the WikiHow and ArXiv/PubMed datasets do not have extractive labels.
Therefore, we follow the same steps as the original PreSumm paper~\citep{liu-lapata-2019-text} to obtain the ground truth extractive labels.
Additionally, due to the large training corpus size for WikiHow and ArXiv/PubMed, we limit the pool of unlabeled examples by randomly sampling a subset of 1,000 articles in each training cycle.
We choose the above three datasets to cover a diverse set of scenarios: TweetSumm contains real-world conversations and human-annotated extractive summaries, WikiHow contains medium-sized documents, and ArXiv/PubMed contains long documents.

\subsection{Training and Evaluation Details.}

\noindent \textbf{Training.} We use the TransformerSum repository\footnote{\url{https://transformersum.readthedocs.io/en/latest/}} to implement our training pipeline.
We use PreSumm as our teacher model $M$ and experiment with two backbones: \texttt{distilbert-base-uncased} and \texttt{bert-base-uncased}.
We perform experiments in two settings: \textbf{1)} data-scarce setting where fix the size of the labeled set $D_l$ to 50 for all the datasets, and \textbf{2)} data-abundant setting where we set the size of $D_l$ to 500.
We set $N_{cycles}$ to 50 for all experiments.
We add 5 pseudo-labels to the training set in each cycle, thus resulting in a final training set size of 300 (50 labeled + 250 pseudo-labeled examples.)
We set the summary size $k$ to 4 for TweetSumm and 8 for WikiHow and ArXiv/PubMed.
We base these summary sizes on the average summary size of the labeled training set.
For training, we start with a learning rate of $2 \times 10^{-5}$ on all the datasets and use a cyclic learning rate scheduler during training~\citep{smith2015cyclical}, which is the default setting in TransformerSum.
Additionally, we use AdamW as our optimizer with $\epsilon = 1 \times 10^{-8}, \beta_1 = 0.9, \beta_2 = 0.99$.
We train all our models on a single V100 GPU with 12G VRAM.
We repeat each experiment for three different seeds and report the mean and standard deviation in our results unless otherwise stated.

\begin{table*}[t]
    \centering
    \resizebox{\linewidth}{!}
    {%
    \begin{tabular}{lccccccccc}
        \toprule
         & \multicolumn{3}{c}{\textbf{TweetSumm}} & \multicolumn{3}{c}{\textbf{WikiHow}} &
         \multicolumn{3}{c}{\textbf{ArXiv/Pubmed}} \\
         \cmidrule(r){2-10}
         \textbf{Method} & \textbf{R-1 (\%)} & \textbf{R-2 (\%)} & \textbf{R-L (\%)} & \textbf{R-1 (\%)} & \textbf{R-2 (\%)} & \textbf{R-L (\%)} & \textbf{R-1 (\%)} & \textbf{R-2 (\%)} & \textbf{R-L (\%)} \\
        \midrule
        \multicolumn{10}{c}{\textbf{DistilBERT$_{base}$ (50 labels)}}\\
        \midrule
        Random & 36.7 (1.5) & 25.4 (1.4) & 36.7 (1.3) & 19.7 (1.4) & 1.5 (1.1) & 7.2 (1.3) & 19.5 (1.3) & 2.9 (0.9) & 7.8 (1.2) \\
         Confidence & 43.5 (1.4) & 35.1 (1.2) & 46.8 (1.1) & 21.3 (0.4) & 3.7(0.8) & 10.3 (1.0) & 23.4 (1.1) & 5.2 (0.7) & 12.5 (1.1) \\
         \quad + G-4 relabeling & {55.4 (1.3)} & \textbf{46.7 (0.6)} & \underline{56.1 (0.9)} & {22.1 (0.4)} & {5.7 (0.6)} & {13.5 (0.7)} & {23.8 (0.8)} & {7.3 (1.3)} & {15.3 (0.8)} \\
         Confidence + G-4 score & 46.8 (1.3) & 37.4 (0.4) & 48.3 (1.2) & 21.7 (0.9) & 4.6 (0.4) & 12.1 (1.1) & 24.1 (0.9) & 6.7 (0.3) & 13.8 (1.4) \\
        \  + G-4 relabeling (Ours) & \textbf{57.6 (1.2)} & \underline{46.3 (1.7)} & \textbf{56.2 (1.3)} & \textbf{22.7 (0.3)} & \textbf{5.9 (0.4)} & \textbf{13.8 (0.5)} & \textbf{24.7 (0.9)} & \textbf{8.1 (1.3)} & \textbf{15.9 (0.8)} \\
        Confidence + L-3 score & 45.7 (1.1) & 36.9 (0.2) & 47.8 (1.2) & 21.6 (0.4) & 4.1 (0.5) & 11.1 (0.8) & 23.9 (0.9) & 6.1 (0.3) & 12.9 (1.3) \\
        \  + L-3 relabeling (Ours) & \underline{56.2 (1.1)} & {45.1 (1.2)} & {55.9 (1.1)} & \underline{22.3 (0.1)} & \underline{5.8 (0.2)} & \underline{13.6 (0.3)} & \underline{24.5 (0.6)} & \underline{7.7 (1.4)} & \underline{15.7 (0.3)} \\
        \midrule
        \multicolumn{10}{c}{\textbf{BERT$_{base}$ (50 labels)}}\\
        \midrule
        TSL (50:500) & 49.0 & 37.7 & 48.2 & - & - & - & - & - & - \\
        Random & 45.4 (1.4) & 32.4 (1.9) & 42.5 (1.8) & 22.1 (1.7) & 2.4 (1.5) & 9.6 (1.5) & 23.3 (1.4) & 6.1 (1.2) & 12.4 (1.3) \\
         Confidence & 49.7 (1.6) & 39.5 (1.4) & 49.4 (1.3) & 24.5 (0.6) & 4.8 (1.1) & 12.8 (1.0) & 27.6 (1.1) & 7.7 (0.7) & 14.2 (1.2) \\
         \quad + G-4 relabeling & {57.8 (1.2)} & {50.3 (0.5)} & {58.9 (1.2)} & \textbf{26.4 (0.3)} & \textbf{7.3 (0.5)} & \textbf{16.4 (0.8)} & {28.7 (0.9)} & {9.5 (1.1)} & {17.1 (0.8)} \\
         Confidence + G-4 score & 52.3 (1.6) & 42.8 (0.7) & 51.0 (1.4) & 25.2 (0.7) & 5.6 (0.5) & 13.1 (1.0) & 27.7 (0.9) & 7.9 (0.2) & 15.5 (1.3) \\
        \ + G-4 relabeling (Ours) & \textbf{58.9 (1.4)} & \textbf{50.4 (0.8)} & \textbf{59.4 (1.5)} & \underline{26.1 (0.4)} & \underline{7.2 (0.6)} & \underline{15.9 (0.9)} & \textbf{29.1 (0.7)} & \textbf{9.7 (1.2)} & \textbf{17.7 (0.6)} \\
        Confidence + L-3 score & 51.7 (1.2) & 41.6 (1.2) & 50.3 (1.2) & 25.9 (0.3) & 5.2 (0.2) & 13.0 (0.8) & 27.6 (0.4) & 7.9 (0.5) & 15.3 (1.1) \\
        \ + L-3 relabeling (Ours) & \underline{58.4 (1.2)} & {50.1 (0.3)} & \underline{59.1 (1.2)} & {26.0 (0.2)} & {6.9 (0.3)} & {15.1 (0.2)} & \underline{29.0 (0.5)} & \underline{9.4 (0.7)} & \underline{17.4 (0.3)} \\
        \midrule
        \multicolumn{10}{c}{\textbf{BERT$_{base}$ (500 labels)}}\\
        \midrule
        TSL (500:500) & 59.0 & 48.3 & 58.2 & - & - & - & - & - & - \\
        Random & 55.1 (1.4) & 42.7 (1.1) & 50.3 (1.2) & 25.6 (1.3) & 4.5 (1.1) & 15.2 (1.3) & 25.4 (1.5) & 9.5 (1.2) & 24.1 (1.2) \\
         Confidence & 61.8 (0.7) & 54.9 (0.8) & 60.3 (0.9) & 28.4 (0.6) & 8.0 (1.1) & 22.5 (1.0) & 29.4 (0.5) & 11.5 (0.6) & 27.7 (0.8) \\
        \ + L-3 score & 63.4 (0.5) & 55.6 (0.8) & 62.1 (0.7) & 28.9 (0.4) & 8.2 (0.5) & 28.4 (0.4) & 31.7 (0.3) & 11.8 (0.2) & 29.4 (0.4) \\
        \ + L-3 relabeling (Ours) & \textbf{64.2 (0.2)} & \textbf{56.2 (0.4)} & \textbf{62.8 (0.6)} & \textbf{30.7 (0.4)} & \textbf{8.8 (0.3)} & \textbf{29.5 (0.3)} & \textbf{33.5 (0.3)} & \textbf{12.3 (0.2)} & \textbf{32.2 (0.3)} \\
        \bottomrule
    \end{tabular}
    }
    \caption{Mean (Std.) ROUGE F-1 scores of different pseudo-labeling strategies.
    R-1, R-2, and R-L denote ROUGE-1, ROUGE-2, and ROUGE-L metrics, respectively.
    TSL results from~\cite{zhuang2023self}.
    Refer to Section~\ref{sec:baselines} for method details.
    \textbf{Bold} indicates the best-performing and  \underline{underline} denotes the second-best performing method, respectively.}
    \label{tab:full_results}
\end{table*}

\noindent \textbf{Evaluation.} First, we use the standard ROUGE-$n$ (R-$n$) and ROUGE-L (R-L) F1 scores~\citep{lin-2004-rouge} on the test set to evaluate summary quality.
ROUGE-$n$ measures the $n$-gram overlap between the predicted and the ground truth summaries, whereas ROUGE-L also takes into consideration the order of $n$-grams.
We use \texttt{pyrouge} to compute ROUGE scores in our setup and report them in Table~\ref{tab:full_results}.
Additionally, we use L-Eval, which is the same as G-Eval~\citep{liu-etal-2023-g} but uses LLaMA instead of GPT-4 as the backend.
We use a metric like G-Eval as it substantially surpasses ROUGE scores and previous model-based evaluation metrics, such as BERTScore~\citep{zhang2019bertscore} and BARTScore~\citep{yuan2021bartscore}, for summarization.
More importantly, it is shown to be better aligned with human preferences than the existing automated evaluation metrics for text summarization.
Finally, G-Eval has been shown to not exhibit bias towards GPT-generated content.
We report L-Eval results in Table~\ref{tab:sup_vs_semisup}.

\subsection{Baselines}
\label{sec:baselines}
We compare our method with the following baselines:
\textbf{1) PreSumm}~\citep{liu-lapata-2019-text}. The original PreSumm model that pretrains a BERT model for summarization.
We train two PreSumm models -- one on a limited training set with 50 labeled examples to match the starting point of our semi-supervised setting and another with 300 labeled examples, the same as the dataset size at the end of our training cycle.
\textbf{2) Teacher-Student Learning (TSL)}~\citep{zhuang2023self}.
Current state-of-the-art semi-supervised method on TweetSumm.
The teacher-student learning framework uses a similar formulation for computing model confidence to ours, as follows: $C_j = \sum_{i=1}^{n} (C_{ij})/n_j$.
Here, $C_{ij} = p_{ij}q_{ij} + (1 - p_{ij})(1 - q_{ij})$, where $p_{ij}$ is the probability of sentence $i$ being selected for summary for dialog $j$ estimated by the teacher model, and $q_{ij} = 1 \ \textrm{if} \ {p_{ij}} \ \textrm{in top 4, else} \ 0$.
We report the performance of the TSL (50:500) and TSL (500:500) models from the paper, as they are the closest to our setup (50/500 labeled examples + 500 unlabeled examples).
\textbf{3) Confidence + G-4 relabeling + G-4 score (Ours).} Our proposed method following the methodology in Section~\ref{sec:method}.
We first use the PreSumm teacher model to shortlist 50 pseudo-labels (Stage 1 and 2), relabel them using GPT-4 (Stage 3), and then select the top 5 using GPT-4 score (Stage 4).
\textbf{4) Confidence + G-4 score.} We skip Stage 3 from \textbf{3)} to directly score the top 50 PreSumm pseudo-labels using GPT-4.
We run this baseline to measure the effect of relabeling in our pipeline.
\textbf{5) Confidence + G-4 relabeling.} We skip Stage 4 from \textbf{3)} and select the final 5 pseudo-labels based on PreSumm confidence.
\textbf{6) Confidence + L-3 relabeling + L-3 score (Ours).} Same as \textbf{3)} but using LLaMA-3.
\textbf{7) Confidence + L-3 score (Ours).} Same as \textbf{4)} but using LLaMA-3.
\textbf{8) Confidence.} We skip Stage 3 and 4 from from \textbf{3)} and select 5 PreSumm pseudo-labels based on PreSumm confidence.
\textbf{9) Random.} Same as \textbf{6)} but instead of using teacher confidence defined in Equation~\ref{eq:conf}, we randomly select five PreSumm pseudo-labels to include in each cycle.

\begin{table}[t!]
    \centering
    \resizebox{\linewidth}{!}
    {%
    \begin{tabular}{lcccccc}
        \toprule
         & \multicolumn{2}{c}{\textbf{TweetSumm}} & \multicolumn{2}{c}{\textbf{WikiHow}} &
         \multicolumn{2}{c}{\textbf{ArXiv/Pubmed}} \\
         \cmidrule(r){2-7}
         \textbf{Method} & \textbf{R-2} & \textbf{L-Eval} & \textbf{R-2} & \textbf{L-Eval}  & \textbf{R-2} & \textbf{L-Eval} \\
        \midrule
        PreSumm (50 labels) & 37.1 (1.1) & 31.2 (0.5) & 3.2 (0.8) & 34.2 (1.5) & 7.3 (0.9) & 13.5 (1.2) \\
        PreSumm (300 labels) & 51.1 (2.1) & {60.5 (1.2)} & 7.6 (0.6) & {68.1 (1.1)} & 10.8 (0.9) & {49.5 (2.4)} \\
        PreSumm (500 labels) & 54.4 (1.2) & {67.1 (0.3)} & 7.9 (0.5) & {74.4 (0.6)} & 11.3 (0.5) & {58.2 (1.1)} \\
        PreSumm (750 labels) & \underline{56.1 (0.7)} & \underline{70.3 (0.5)} & \underline{8.5 (0.4)} & \underline{76.5 (0.4)} & \underline{12.1 (0.7)} & \underline{62.8 (0.7)} \\
        \midrule
        \multicolumn{7}{c}{\textbf{50 labels}} \\
        \midrule
        Random & 32.4 (1.9) & 32.1 (1.1) & 2.4 (1.5) & 37.7 (1.6) & 6.1 (0.2) & 15.1 (2.3) \\
         Confidence + G-4 score & 42.8 (0.7) & 46.2 (0.2) & 5.6 (0.5) & 59.4 (1.3) & 7.9 (0.2) & 40.1 (1.9) \\
        \ + G-4 relabeling (Ours) & {50.4 (0.8)} & {58.4 (0.4)} & {7.2 (0.6)} & {70.3 (1.4)} & {9.7 (1.2)} & {52.5 (1.3)} \\
        Confidence + L-3 score & 41.6 (1.2) & 45.8 (0.7) & 5.2 (0.2) & 57.5 (1.4) & 7.9 (0.5) & 37.1 (1.8) \\
        \ + L-3 relabeling (Ours) & {50.1 (0.3)} & {56.3 (0.9)} & {6.9 (0.3)} & {69.3 (2.1)} & {9.4 (0.7)} & {49.3 (1.4)} \\
        \midrule
        \multicolumn{7}{c}{\textbf{500 labels}} \\
        \midrule
        Random & 42.7 (1.1) & 52.3 (1.2) & 4.5 (1.1) & 52.7 (1.8) & 9.5 (1.2) & 44.1 (0.9) \\
        Confidence + L-3 score & 55.6 (0.8) & 69.2 (0.7) & 8.2 (0.5) & 75.2 (1.4) & 11.8 (0.2) & 60.2 (0.5) \\
        \ + L-3 relabeling (Ours) & \textbf{56.2 (0.8)} & \textbf{71.2 (0.9)} & \textbf{8.8 (0.3)} & \textbf{77.3 (1.3)} & \textbf{12.3 (0.2)} & \textbf{65.7 (0.3)} \\
        \bottomrule
    \end{tabular}
    }
    \caption{\textbf{Fully-supervised methods (first two rows) semi-supervised approaches (remaining rows).} All models use BERT$_{base}$ as PreSumm's backbone. The number of labeled examples for fully supervised models is shown in brackets.
    The semi-supervised methods use 50/500 labeled and 250 unlabeled examples.}
    \label{tab:sup_vs_semisup}
\end{table}

\section{Results}
\label{sec:results}
We now discuss results in  Table~\ref{tab:full_results} and Table~\ref{tab:sup_vs_semisup}.
Table~\ref{tab:full_results} reports the ROUGE scores for different pseudo-labeling strategies for the three datasets.
All the methods operate in (50:250) or (500:250) setups, except TSL (50:500), which uses 500 unlabeled examples.
Notably, Table~\ref{tab:full_results} has two blocks of results, one for DistilBERT$_{base}$ as the backbone of the PreSumm model and another for BERT$_{base}$ as the PreSumm backbone.
Table~\ref{tab:sup_vs_semisup} compares L-Eval scores of our semi-supervised approach with fully-supervised setups.

\begin{figure}[htbp]
    \centering
    \includegraphics[width=\linewidth]{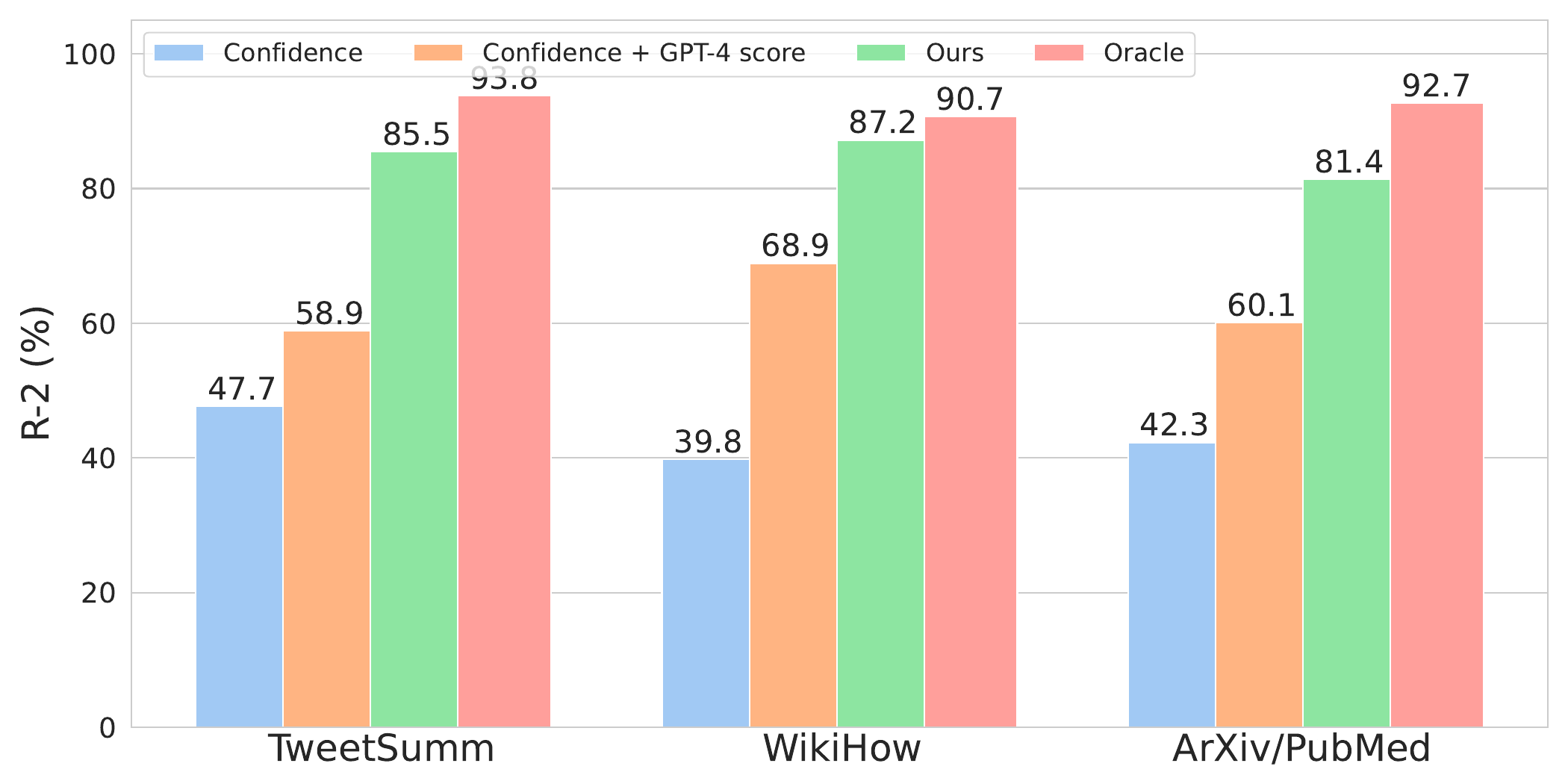}  
    \caption{\textbf{Quality of pseudo-labels by different strategies (data-scarce setup).} The y-axis denotes the ROUGE-2 scores of the top 5 pseudo-labels computed against the respective ground truths. All results are for BERT$_{base}$ as the backbone for PreSumm and three random seeds. Refer to Section~\ref{sec:pseudo-quality} for complete details.}
    \label{fig:pseudoquality}
\end{figure}

\begin{figure*}[t]
    \centering
    \includegraphics[width=\linewidth]{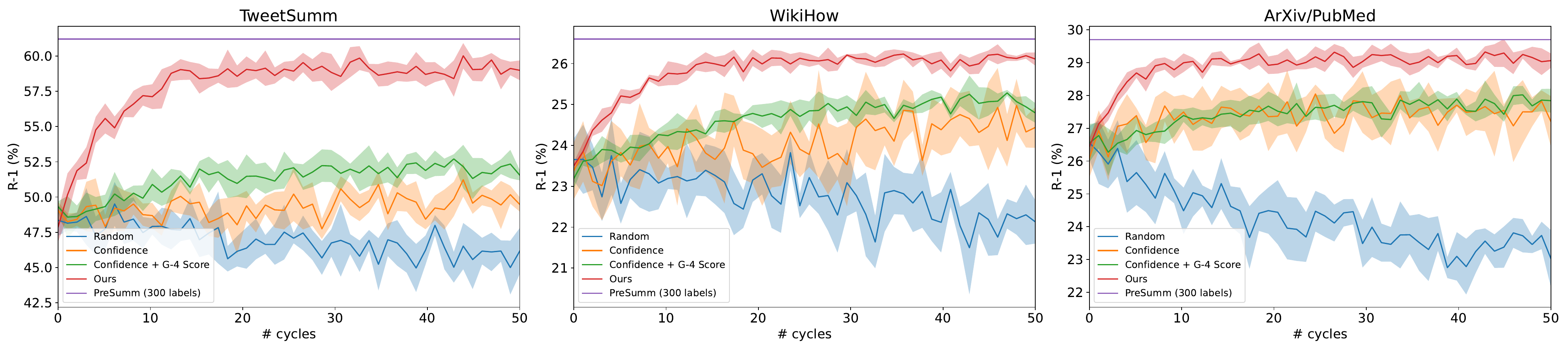}
    \caption{\textbf{ROUGE-1 curves v/s \# cycles for data-scarce setting.} Each cycle denotes an addition of 5 new pseudo-labels to the training set. All results use BERT$_{base}$ as the backbone for PreSumm. The curves are averaged for three seeds (the width denotes the std). Note that we report the GPT-4 version of our method here.}
    \label{fig:r2_graph}
\end{figure*}

\subsection{Comparison of Pseudo-label Selection Strategies}
\label{sec:pseudo-quality}
Referring to Table~\ref{tab:full_results}, we note that all pseudo-label selection strategies outperform the random baseline.
The ``Random" baseline performs worse than the fully supervised counterpart on all datasets (R-2 in Table~\ref{tab:full_results} v/s R-2 in Table~\ref{tab:sup_vs_semisup}), meaning that the \textit{majority} of the shortlisted PreSumm pseudo-labels are low-quality.
Using teacher confidence leads to slight performance gains on all the datasets, and adding GPT-4 score further improves the results (``Confidence" v/s ``Confidence + G-4 score" in Table~\ref{tab:full_results}).
These improvements indicate that the shortlisted PreSumm pseudo-labels include some good-quality pseudo-labels, too, and using GPT-4 to rate those pseudo-labels is crucial to picking them.
We see similar trends when using LLaMA-3.

To further confirm our findings, we conduct a qualitative study in the data-scarce setup, where we compute the ROUGE scores of the final 5 pseudo-labels for each method against the respective ground truth summaries, and Figure~\ref{fig:pseudoquality} shows the mean ROUGE-2 of the five selected pseudo-labels.
To clarify, we obtained the ``Oracle" results by directly selecting the final 5 pseudo-labels using ROUGE-2 scores computed against the ground truth.
We note a stark difference between ``Confidence" and ``Oracle," which shows that relying solely on teacher confidence consistently leads to a selection of low-quality pseudo-labels. 
Combining GPT-4 score with teacher confidence is effective (``Confidence + G-4 score"), and adding the GPT-4 relabeling greatly boosts the quality of selected pseudo-labels (``Ours").

\subsection{Effect of Relabeling}
Referring to Tables~\ref{tab:full_results} and ~\ref{tab:sup_vs_semisup}, we observe that relabeling with LLMs leads to a significant boost in the summarization performance in terms of both ROUGE scores and L-Eval.
When using BERT$_{base}$ as the backbone, we note that the ROUGE-1 improves from 52.3 to 58.9 on the TweetSumm dataset, 25.2 to 26.1 on the WikiHow dataset, and 27.7 to 29.1 on the ArXiv/PubMed dataset.
GPT-4 relabeling is also effective when using teacher confidence without GPT-4 score (``Confidence" v/s ``Confidence + G-4 relabeling").
Our previous qualitative study also supports these results, showing that relabeling improves the quality of pseudo-labels.
We observe similar trends when using DistilBERT$_{base}$ as PreSumm's backbone and LLaMA-3 instead of GPT-4.
When using 500 labels, we note boosts in performance but the relative scale is smaller compared to when using 50 labels.

We conduct additional testing to analyze the performance of our best- and second-best-performing models, both of which involve relabeling.
We find that the p-value < 0.016 for Welch’s test for an R-1 of 58.9 (1.4) for ``Confidence + G-4 score + G-4 relabelling" v/s 57.8 (1.2) for ``Confidence + G-4 relabeling" on the TweetSumm dataset, denoting the differences are significant.

\subsection{On Sample Efficiency of Different Approaches}
Referring to Table~\ref{tab:sup_vs_semisup}, we now analyze the sample efficiency of different learning algorithms.
For fully supervised methods, we note that including more labeled examples improves L-Eval and ROUGE scores across the board (``PreSumm (50 labels)" v/s ``PreSumm (300 labels)").
Our semi-supervised approach using 50 labels with GPT-4 relabeling and GPT-4 score achieves competitive performance to the fully supervised PreSumm model trained on 300 labels.
Notably, we get better L-Eval scores than ``PreSumm (300 labels)" on WikiHow and ArXiv/PubMed datasets and are competitive on TweetSumm.
This is encouraging, as ``PreSumm (300 labels)" approximates the best-case scenario of 100\% high-quality labels in the training set.
In the data-abundant setting, our proposed method with LLaMA outperforms the respective fully supervised model in terms of both ROUGE and L-Eval.

We plot the R-1 scores against the number of training cycles, as shown in
Figure~\ref{fig:r2_graph}. 
Overall, ``Random" setting is highly unstable, ``Confidence + G-4 score" slightly improves over ``Confidence" on TweetSumm and WikiHow, but more importantly, it is consistently more stable.
Finally, our method with GPT-4 scoring and relabeling not only significantly boosts the R-1 scores (visible gap between ``Ours" and the rest) but also does so at a much faster rate.
For all the datasets, our method peaks and stabilizes under 20 cycles (100 pseudo-labels), further endorsing the sample efficiency of our method compared to other approaches.

\section{Conclusion}
To conclude, we propose a novel prompt-based pseudo-labeling strategy for semi-supervised extractive text summarization.
Our method effectively combines teacher confidence and knowledge of LLMs like GPT-4 to improve pseudo-labeling in low-resource summarization setups.
Adding a GPT-4 critic to relabel and rank pseudo-labels leads to highly performant summarization models that are competitive with their fully supervised counterparts while utilizing 6$\times$ fewer labeled examples.

\section*{Limitations}
Our results indicate a promising potential to use LLMs for semi-supervised extractive text summarization;
however, this work has a few limitations, as detailed in this section.

First, we rely completely on LLMs pseudo-labels for scoring and relabelling, which is not perfect and we do not currently have a way to automatically judge them. We need a human in the loop to judge the quality of such pseudo-labels.

Second, we can also potentially improve the sample efficiency of the model.
Currently, we relabel all the top 50 pseudo-labels selected by PreSumm in Stage 3.
This is cheaper than pseudo-labeling all the pseudo-labels, but there's a chance that LLM relabels an already good-quality summary, as we note from our analysis that the initial shortlist contains some good pseudo-labels.
Therefore, we can develop an additional strategy to decide whether to relabel a given pseudo-label.
Such a method would further reduce costs for using closed-source models like GPT-4.
We hope this work furthers research toward incorporating LLMs to address such problems for low-resource semi-supervised tasks.

\section*{Ethics Statement}
We use GPT and LLaMA-3 to generate new examples, and even though they is instruction-tuned, some generations might depict undesirable biases with respect to the current objective.
For the stated reason, we recommend using our proposed models with human monitoring when employing the method for real-world settings. 
Practitioners may also consider explicitly debiasing language models for their specific use cases~\citep{barikeri2021redditbias,schick2021self}.

\bibliography{custom}




\end{document}